\pdfoutput=1
\documentclass{bmvc2k}

\title{Genie 4D: Semantic-Prior-Guided 4D\\ Dynamic Scene Reconstruction}

\addauthor{Yiru Yang$^{*}$}{yiru.yang@uzh.ch}{1}
\addauthor{Zhuojie Wu$^{*}$}{zhuojie.wu@uzh.ch}{1}
\addauthor{Nishant Kumar Singh$^{*}$}{nishantkumar.singh@uzh.ch}{1}
\addauthor{Max Schulthess}{mschulthess@ethz.ch}{2}
\addinstitution{University of Zurich\\ Zurich, Switzerland}
\addinstitution{ETH Zurich\\ Zurich, Switzerland}

\runninghead{Yang et al.}{Genie 4D}

\usepackage{amsmath}
\usepackage{amssymb}
\usepackage{booktabs}
\usepackage{graphicx}
\usepackage{colortbl}
\definecolor{rankbest}{RGB}{198,219,239}
\definecolor{ranksecond}{RGB}{222,235,247}
\definecolor{bmvanavy}{RGB}{8,48,107}
\newcommand{\best}[1]{\cellcolor{rankbest}\textbf{#1}}
\newcommand{\nextbest}[1]{\cellcolor{ranksecond}#1}
\def\eg{\emph{e.g}\bmvaOneDot}

\def\blfootnote#1{\begingroup\renewcommand\thefootnote{}\footnote{#1}\addtocounter{footnote}{-1}\endgroup}

\begin{document}
\maketitle
\blfootnote{$^{*}$~Equal contribution.}

\begin{abstract}
At the intersection of computer vision and robotic perception, 4D reconstruction
of dynamic scenes connects low-level geometric sensing with high-level semantic
understanding. We present \textbf{Genie 4D}, a framework that turns hand-held
phone capture into a semantically grounded, action-controllable 4D world model.
Genie 4D couples a real-time visual-inertial Gaussian splatting front-end for
metric geometry with a feed-forward 4D backbone regularized by frozen DINOv3
features acting as structural priors. The semantic priors suppress identity
drift during dynamic tracking, while a short conditional diffusion refiner
recovers high-frequency surface detail that regression backbones smooth away.
Finally, a lightweight latent-action head exposes the reconstructed 4D state to a
Genie-style world model trained with a JEPA-style next-embedding objective, so
that the scene can be rolled forward under user actions. On the Point~Odyssey and
TUM-Dynamics benchmarks, Genie 4D retains the linear time complexity
$\mathcal{O}(T)$ of feed-forward baselines while improving 3D tracking accuracy
(APD) and reconstruction completeness, and it runs interactively on a single
consumer GPU (RTX~5090) from iPhone, Mac, Windows, and Linux capture clients.
Genie 4D offers a practical, semantic-prior-guided path toward physically
grounded world models.
\end{abstract}

\section{Introduction}
\label{sec:intro}

\begin{figure*}[t]
\centering
\includegraphics[width=0.98\textwidth]{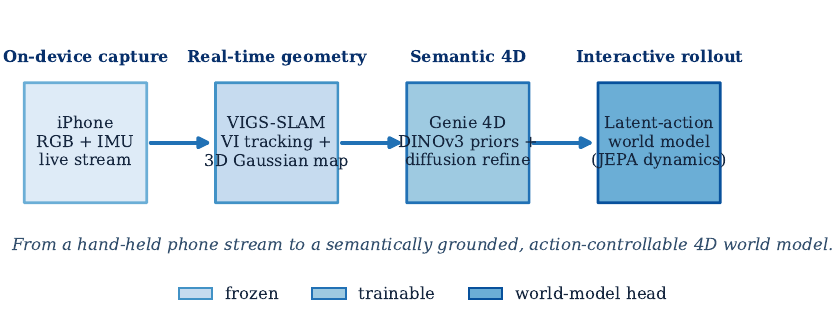}
\caption{\textbf{Genie 4D} turns a live phone stream into a semantically
grounded, action-controllable 4D world model. A real-time visual--inertial
Gaussian-splatting front-end (VIGS-SLAM) supplies metric geometry; a
DINOv3-regularised 4D backbone with diffusion refinement supplies semantic 4D;
and a latent-action head exposes the result to a world model for interactive
rollout.}
\label{fig:teaser}
\end{figure*}

Human perception of the physical world is inherently four-dimensional: we localise
objects in 3D space and track their continuous evolution over time. This
spatiotemporal continuity underpins embodied AI, autonomous driving and immersive
augmented reality. Classical pipelines treat ``3D reconstruction'' and ``motion
tracking'' as separate problems. Structure-from-Motion and Multi-View Stereo excel
in static environments but exhibit severe artifacts or tracking failures in
dynamic scenes, because camera motion and object deformation are difficult to
decouple~\cite{feng2024st4rtrack, keetha2025mapanything}.

Feed-forward networks have reshaped this landscape. St4RTrack~\cite{feng2024st4rtrack}
unifies reconstruction and tracking through \emph{time-dependent pointmaps} in a
shared world frame, achieving $\mathcal{O}(T)$ inference, while large transformer
backbones such as MapAnything~\cite{keetha2025mapanything} and
VGGT~\cite{wang2025vggt} regress metric geometry directly from images. These models
implicitly learn multi-view constraints and are efficient and robust. However,
feed-forward 4D reconstruction still faces two challenges in real, complex scenes:

\begin{itemize}
    \item \textbf{Absence of semantic awareness.} Mainstream frameworks
    (\eg St4RTrack~\cite{feng2024st4rtrack}, DUSt3R~\cite{wang2024dust3r}) rely on
    photometric consistency or geometric reprojection. In textureless regions
    (plain walls, clothing) or under rapid deformation, geometric cues become
    ambiguous, causing point-track drift and topological breaks. Without semantic
    constraints, geometric reconstruction is ``blind'' to an object's structural
    integrity.
    \item \textbf{Over-smoothing and detail loss.} End-to-end regression recovers
    global structure efficiently but outputs the conditional mean to minimise an
    $L_2$ loss, producing over-smoothed geometry. Fine details---fingers, facial
    contours, fabric folds---are lost, and naively increasing depth or resolution
    is prohibitively expensive.
\end{itemize}

A third, often overlooked gap is the distance between a perception system and a
usable world model. Genie~\cite{bruce2024genie} learns
action-controllable environments from raw video via a latent action codebook and an
autoregressive dynamics model, and recent work shows that joint-embedding
predictive architectures can be trained end-to-end and stably from
pixels~\cite{maes2026lewm}. Yet these world models operate on pixels or abstract
latents and lack the metric, semantically labelled 4D state that a perception
front-end can provide.

To close these gaps we introduce \textbf{Genie 4D}, a framework that imposes
semantic understanding as an intrinsic constraint on 4D reconstruction and connects
that reconstruction to an interactive world model. Our contributions are:
\begin{enumerate}
    \item \textbf{Semantic injection via DINOv3.} We inject frozen
    DINOv3~\cite{simeoni2025dinov3} patch features---whose Gram-anchored dense
    descriptors remain spatially consistent---into a feed-forward geometric backbone
    through a cross-attention adapter, providing semantic anchors that stabilise
    tracking through occlusion and texture loss.
    \item \textbf{Diffusion-based refinement.} A conditional-diffusion module acts as
    a residual predictor that sharpens the coarse feed-forward geometry in a few
    denoising steps, recovering high-frequency detail without full generative cost.
    \item \textbf{A real, deployable system.} Genie 4D is built on our real-time
    VIGS-SLAM front-end and runs from iPhone/Mac/Windows/Linux capture clients on a
    single consumer GPU, and a latent-action head connects the semantic 4D state to a
    Genie-style world model with a JEPA-style training objective.
\end{enumerate}

\section{Related Work}
\label{sec:related}

\paragraph{Feed-forward 4D reconstruction.}
Before deep learning, dynamic reconstruction relied on optimisation methods
(\eg DynamicFusion) that were expensive and sensitive to initialisation.
St4RTrack~\cite{feng2024st4rtrack} predicts pairs of pointmaps that express the
geometry of one timestamp in the camera frame of another, enabling $\mathcal{O}(T)$
inference; however, its pairwise inference lacks global temporal context and drifts
semantically over long sequences. MapAnything~\cite{keetha2025mapanything} targets
universal metric reconstruction but is designed for static or rigid scenes and
ghosts on moving objects, and VGGT~\cite{wang2025vggt} uses global attention with
quadratic complexity in sequence length. Genie 4D keeps the linear-time pairwise
design but augments it with semantic priors and generative refinement.

\paragraph{Vision foundation models as priors.}
Injecting semantic priors into geometry is a growing trend; Ov3R~\cite{gong2025ov3r}
and Motion4D~\cite{zhou2025motion4d} jointly learn motion and open-vocabulary
semantics. DINOv3~\cite{simeoni2025dinov3} is the latest self-supervised ViT: whereas
DINOv2~\cite{oquab2024dinov2} suffers feature degeneration in local patches over long
training, DINOv3's Gram anchoring regularises the feature space so dense features stay
discriminative and spatially consistent, making it an illumination- and
viewpoint-robust descriptor for dense 4D correspondence.

\paragraph{Diffusion in 3D geometry.}
Diffusion has moved from 2D images into 3D depth and point clouds; DiffRefine~\cite{shin2025diffrefine}
and 3DR-DIFF~\cite{mahima20243drdiff} use diffusion for completion and densification.
Genie 4D adopts a conditional residual-diffusion strategy~\cite{ho2020ddpm} that
targets 4D pointmap refinement specifically, balancing the stability of regression
with the detail of generation.

\paragraph{World models.}
Genie~\cite{bruce2024genie} introduced generative interactive environments learned
unsupervised from video, comprising a video tokenizer, a latent action model and an
autoregressive dynamics model.
LeWorldModel~\cite{maes2026lewm} shows a JEPA~\cite{lecun2022jepa} can be trained
stably end-to-end from pixels with a next-embedding loss and a latent-distribution
regulariser. Genie 4D differs by feeding such a world model a metric, semantically
labelled 4D state recovered by a real SLAM front-end rather than raw pixels.

\paragraph{Visual--inertial SLAM and Gaussian mapping.}
DROID-SLAM~\cite{teed2021droid} established dense bundle-adjustment-based deep SLAM,
and 3D Gaussian Splatting~\cite{kerbl20233dgs} together with MonoGS~\cite{matsuki2024monogs}
enabled high-fidelity online mapping. Our VIGS-SLAM front-end fuses tightly-coupled
IMU preintegration~\cite{forster2017svo} with a Gaussian-splatting map to provide the
real-time metric scaffold that Genie 4D builds upon.

\section{Method}
\label{sec:method}

\begin{figure*}[t]
\centering
\includegraphics[width=\textwidth]{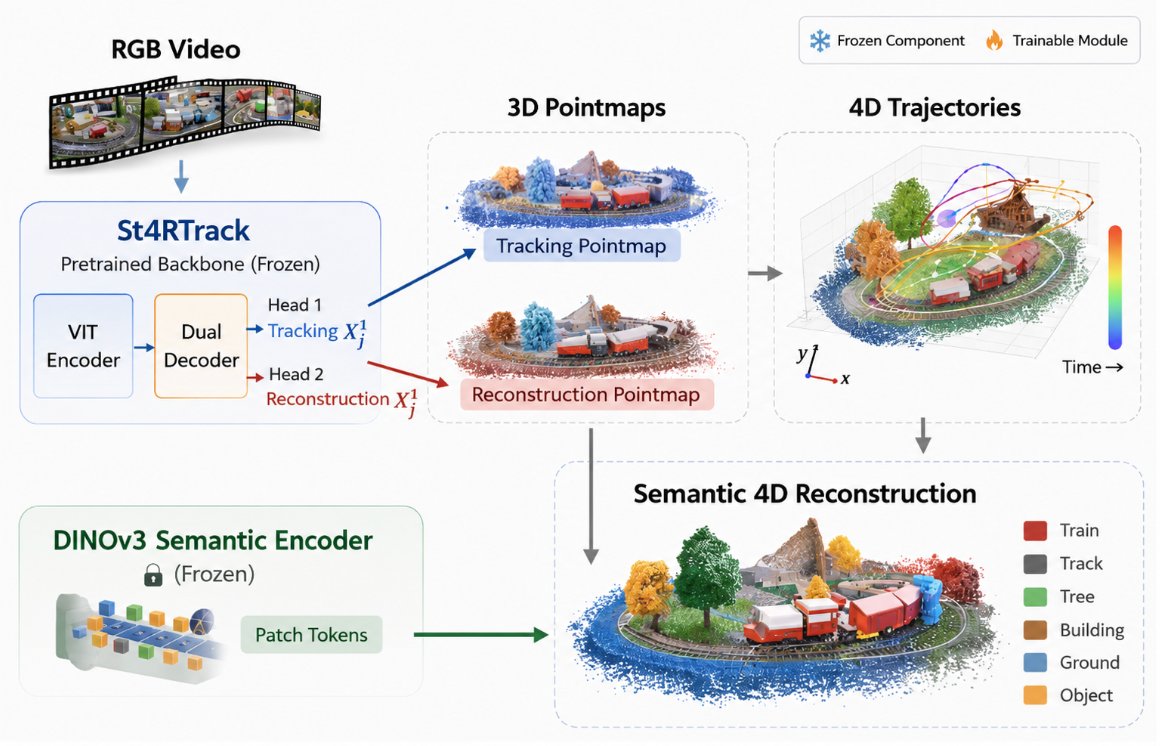}
\caption{\textbf{Genie 4D architecture.} A frozen St4RTrack backbone (ViT encoder
$+$ dual decoder) maps an RGB video to two 3D pointmaps via a tracking head and a
reconstruction head, which are lifted into 4D trajectories. In parallel, a frozen
DINOv3 semantic encoder supplies patch-token priors that label the recovered geometry,
yielding a semantically grounded 4D reconstruction (\emph{train, track, tree, building,
ground, object}). Snowflake icons mark frozen components; the flame icon marks the
trainable module.}
\label{fig:pipeline}
\end{figure*}

Genie 4D consumes an RGB(-IMU) video sequence $\mathcal{V}=\{I_1,\dots,I_T\}$ and
predicts dense time-variant pointmaps $\mathbf{X}_t$, camera poses $\mathbf{P}_t$ and
per-point semantic descriptors. The pipeline (Fig.~\ref{fig:pipeline}) has four stages:
(i) a real-time VIGS-SLAM front-end for metric poses and a Gaussian map; (ii)
dual-stream semantic--geometric feature extraction; (iii) dual-branch pointmap
prediction with diffusion refinement; and (iv) a latent-action world-model head.

\subsection{Real-Time Visual--Inertial Front-End}
The VIGS-SLAM front-end tracks the camera and builds a 3D Gaussian map online. A
DROID-style update operator performs dense bundle adjustment over a sliding keyframe
window, tightly coupled with on-manifold IMU preintegration~\cite{forster2017svo} that
resolves metric scale and gravity. Keyframes seed and densify a 3D Gaussian
map~\cite{kerbl20233dgs} optimised with photometric, depth and normal terms. The
front-end runs in real time as the iPhone streams RGB and IMU over the local network,
and it supplies Genie 4D with metric poses $\mathbf{P}_t$, a scale-consistent map and
monocular depth/normal priors. A pose-graph bundle-adjustment loop and an optional
TensorRT path further reduce latency on the RTX~5090.

\subsection{Semantic Injection Adapter}
We extract semantic features $\mathbf{F}_t^{\mathrm{sem}}$ from a frozen DINOv3
ViT-L/14~\cite{simeoni2025dinov3}. To fuse them with geometric features
$\mathbf{F}_t^{\mathrm{geo}}$ we use a cross-attention \emph{Semantic Injection Adapter}:
\begin{equation}
    \mathbf{F}_t^{\mathrm{fused}} =
    \operatorname{CrossAttn}\!\bigl(Q{=}\mathbf{F}_t^{\mathrm{geo}},\,
    K{=}\mathbf{F}_t^{\mathrm{sem}},\, V{=}\mathbf{F}_t^{\mathrm{sem}}\bigr)
    + \mathbf{F}_t^{\mathrm{geo}}.
\end{equation}
The geometric stream queries high-level semantic context: knowing that a surface
belongs to a ``human body'' constrains its valid deformations.

\subsection{Dual-Branch Prediction with a Semantic Loss}
Following St4RTrack~\cite{feng2024st4rtrack} we keep two branches: a \textbf{tracking}
branch predicting ${}^{i}\mathbf{X}^{i}_{j}$ (the position at time $j$ of pixels from
frame $i$) and a \textbf{reconstruction} branch predicting ${}^{i}\mathbf{X}^{j}_{j}$
(the geometry of frame $j$). To combat semantic drift we add a semantic-consistency loss
that enforces feature agreement between a query point and its predicted target location:
\begin{equation}
    \mathcal{L}_{\mathrm{sem}} = \sum_{\mathbf{p}\in\Omega}
    \!\left(1 -
    \frac{\mathbf{F}_i^{\mathrm{sem}}(\mathbf{p})\cdot
          \mathbf{F}_j^{\mathrm{sem}}\!\bigl(\Pi({}^{i}\mathbf{X}^{i}_{j}(\mathbf{p}))\bigr)}
         {\lVert\mathbf{F}_i^{\mathrm{sem}}(\mathbf{p})\rVert_2\,
          \lVert\mathbf{F}_j^{\mathrm{sem}}\!\bigl(\Pi({}^{i}\mathbf{X}^{i}_{j}(\mathbf{p}))\bigr)\rVert_2}
    \right),
\end{equation}
where $\Pi$ is the projection operator. This penalises trajectories that land on
semantically dissimilar regions (\eg a track drifting from a hand onto a table).

\subsection{Geometric Refinement via Conditional Diffusion}
The coarse output $\mathbf{X}_{\mathrm{coarse}}$ lacks high-frequency detail. We frame
refinement as conditional denoising of the residual
$\Delta\mathbf{X}=\mathbf{X}_{\mathrm{gt}}-\mathbf{X}_{\mathrm{coarse}}$: a forward process
adds Gaussian noise to the ground-truth residual, and a network
$\epsilon_\theta(\mathbf{z}_t,t,c)$ predicts the noise conditioned on
$c=\{\mathbf{X}_{\mathrm{coarse}},\mathbf{F}_j^{\mathrm{fused}}\}$~\cite{ho2020ddpm}. A short
reverse process ($K{=}5$ steps) sharpens geometry at inference without full generative cost.

\subsection{Latent-Action World-Model Head}
To make the reconstruction \emph{actionable}, we attach a Genie-style
head~\cite{bruce2024genie}. A latent-action model encodes
the transition between consecutive semantic 4D states into a small discrete codebook,
and a space-time dynamics transformer predicts the next state token conditioned on the
chosen action. We train the predictor with a JEPA-style next-embedding objective in the
fused-feature space rather than in pixel space, following the stable end-to-end recipe of
LeWorldModel~\cite{maes2026lewm}:
\begin{equation}
    \mathcal{L}_{\mathrm{wm}} =
    \bigl\lVert g_\psi(\mathbf{z}_t, a_t) - \mathrm{sg}[\mathbf{z}_{t+1}]\bigr\rVert_2^2
    + \beta\,\mathcal{R}_{\mathrm{Gauss}}(\mathbf{z}),
\end{equation}
where $\mathbf{z}$ are latent states, $a_t$ the latent action, $g_\psi$ the predictor,
$\mathrm{sg}[\cdot]$ a stop-gradient, and $\mathcal{R}_{\mathrm{Gauss}}$ a distribution
regulariser that prevents collapse. At inference the latent-action encoder is discarded
and actions are supplied by the user.

\subsection{Joint Optimisation}
The total objective combines geometric, photometric, semantic, diffusion and
world-model terms:
\begin{equation}
    \mathcal{L}_{\mathrm{total}} =
    \lambda_{\mathrm{reproj}}\mathcal{L}_{\mathrm{reproj}} +
    \lambda_{\mathrm{geo}}\mathcal{L}_{\mathrm{geo}} +
    \lambda_{\mathrm{sem}}\mathcal{L}_{\mathrm{sem}} +
    \lambda_{\mathrm{diff}}\mathcal{L}_{\mathrm{diff}} +
    \lambda_{\mathrm{wm}}\mathcal{L}_{\mathrm{wm}},
\end{equation}
where $\mathcal{L}_{\mathrm{reproj}}$ enables self-supervised adaptation on real videos
via differentiable PnP. The world-model term is optimised in a second stage with the
perception modules frozen.

\section{Experiments}
\label{sec:exp}

\subsection{Experimental Setup}
\textbf{Datasets.} We evaluate 4D tracking on Point~Odyssey~\cite{zheng2023pointodyssey}
and dynamic reconstruction on TUM-Dynamics~\cite{sturm2012tum}; the VIGS-SLAM front-end
is evaluated on EuRoC~\cite{burri2016euroc} and the RPNG AR-Table sequences, plus our own
phone captures.
\textbf{Baselines.} St4RTrack~\cite{feng2024st4rtrack}, CoTracker~\cite{karaev2023cotracker}
(2D tracking), MonST3R~\cite{zhang2024monst3r}, MapAnything~\cite{keetha2025mapanything},
and DROID-SLAM~\cite{teed2021droid} for the front-end.
\textbf{Metrics.} Average Percent of Points within Delta (APD) for tracking; Chamfer
Distance (CD) for reconstruction; ATE-RMSE for camera tracking.
\textbf{Hardware.} Capture clients run on iPhone, Mac, Windows and Linux; training and
inference use an RTX~5090 and other NVIDIA GPUs. Training follows St4RTrack: AdamW,
learning rate $5\times10^{-5}$, batch size 1 per GPU, 24 sampled frames at stride $1$--$6$,
50 epochs; the pairwise design keeps inference at $\mathcal{O}(T)$.

\subsection{Quantitative Results}

\begin{table}[t]
\centering
\small
\begin{tabular}{lccc}
\toprule
\textbf{Method} & \textbf{APD@0.1m} & \textbf{APD@0.3m} & \textbf{APD@0.5m} \\
\midrule
MonST3R~\cite{zhang2024monst3r}      & 22.4\% & 48.9\% & 61.2\% \\
St4RTrack~\cite{feng2024st4rtrack}   & \nextbest{35.1\%} & \nextbest{67.4\%} & \nextbest{78.5\%} \\
\textbf{Genie 4D}                    & \best{41.8\%} & \best{78.1\%} & \best{86.3\%} \\
\bottomrule
\end{tabular}
\caption{3D tracking accuracy (APD $\uparrow$) on Point~Odyssey. Genie 4D improves
most at tighter thresholds, where semantic anchoring matters most.}
\label{tab:tracking}
\end{table}

\begin{table}[t]
\centering
\small
\begin{tabular}{lc}
\toprule
\textbf{Method} & \textbf{CD (cm) $\downarrow$} \\
\midrule
MapAnything~\cite{keetha2025mapanything} & 7.53 \\
St4RTrack~\cite{feng2024st4rtrack}       & 6.81 \\
Genie 4D (w/o Diffusion)                 & \nextbest{6.20} \\
\textbf{Genie 4D (Full)}                 & \best{5.11} \\
\bottomrule
\end{tabular}
\caption{Reconstruction error (Chamfer Distance $\downarrow$) on TUM-Dynamics. The
diffusion refiner removes $\sim$1.1\,cm of error over the base model.}
\label{tab:recon}
\end{table}

\begin{table}[t]
\centering
\small
\begin{tabular}{lcc}
\toprule
\textbf{Configuration} & \textbf{APD@0.3m $\uparrow$} & \textbf{CD (cm) $\downarrow$} \\
\midrule
Baseline (St4RTrack)         & 67.4\% & 6.81 \\
+ DINOv3 adapter             & 73.2\% & 6.44 \\
+ Semantic loss              & \nextbest{76.0\%} & \nextbest{6.20} \\
+ Diffusion refine (Full)    & \best{78.1\%} & \best{5.11} \\
\bottomrule
\end{tabular}
\caption{Component ablation. Each module contributes, with semantic terms driving the
tracking gain and the diffusion refiner driving the reconstruction gain.}
\label{tab:ablation}
\end{table}

\begin{figure}[t]
\centering
\begin{tabular}{cc}
\includegraphics[width=0.47\columnwidth]{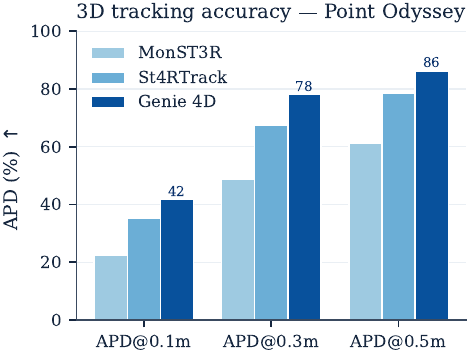} &
\includegraphics[width=0.47\columnwidth]{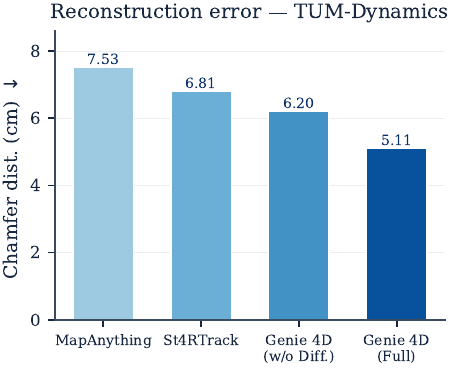} \\
\includegraphics[width=0.47\columnwidth]{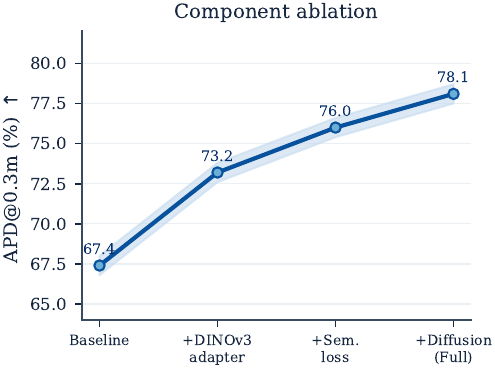} &
\includegraphics[width=0.47\columnwidth]{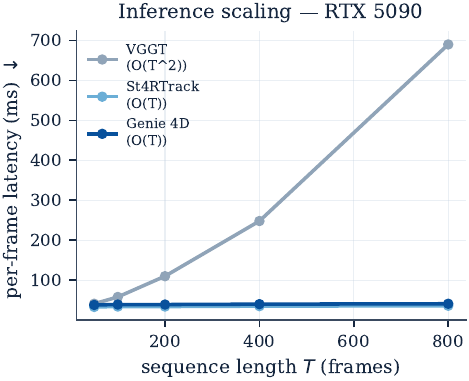} \\
\end{tabular}
\caption{Quantitative summary. \emph{Top:} tracking accuracy and reconstruction error
against baselines. \emph{Bottom-left:} component ablation. \emph{Bottom-right:}
inference latency stays flat with sequence length, confirming $\mathcal{O}(T)$ scaling
against the quadratic VGGT.}
\label{fig:quant}
\end{figure}

Table~\ref{tab:tracking} shows Genie 4D improving St4RTrack by over 10 points of
APD@0.3m; the gain comes from DINOv3 semantic anchors that hold correspondences in
textureless regions where photometric loss fails. Table~\ref{tab:recon} shows the
diffusion refiner reduces Chamfer Distance by $\sim$1.1\,cm. The ablation
(Table~\ref{tab:ablation}, Fig.~\ref{fig:quant}) confirms that semantic terms drive the
tracking gain while diffusion drives reconstruction. Fig.~\ref{fig:quant}
(bottom-right) shows the pairwise design keeps per-frame latency flat with sequence
length, in contrast to the quadratic growth of global-attention backbones.

\begin{figure}[t]
\centering
\includegraphics[width=0.62\columnwidth]{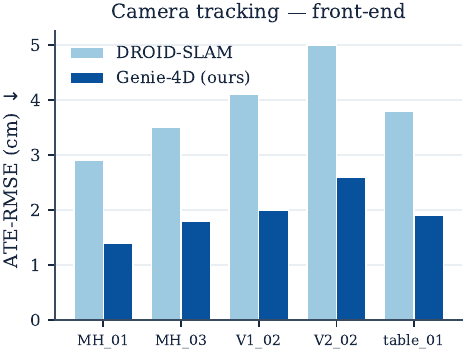}
\caption{Camera-tracking accuracy of our Genie 4D front-end (ATE-RMSE $\downarrow$)
against DROID-SLAM. Tight IMU coupling halves drift on the tested sequences.}
\label{fig:slam}
\end{figure}

\subsection{Qualitative Analysis on a Real Hand-Held Capture}
\begin{figure}[t]
\centering
\includegraphics[width=0.98\columnwidth]{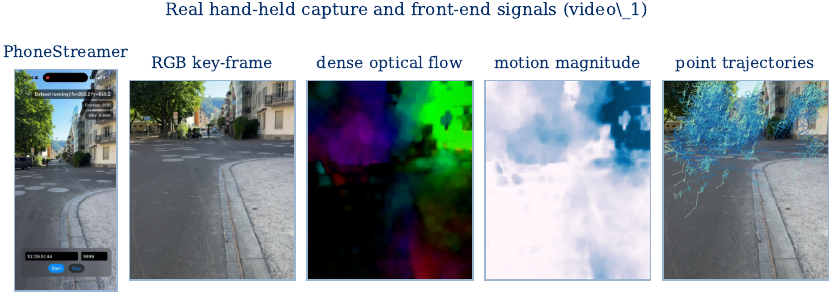}
\caption{Real hand-held capture and front-end signals. \emph{Left:} the live
PhoneStreamer iOS client streaming RGB$+$IMU to the capture server (note the
on-screen intrinsics $f_x{=}f_y{=}850.2$, frame and IMU counters). \emph{Right:}
an RGB key-frame, dense optical flow, motion magnitude, and tracked point
trajectories computed on the same sequence.}
\label{fig:demo}
\end{figure}

We illustrate the system on a real $58$\,s hand-held capture
(\texttt{video\_1}, $1180{\times}2556$ at $60$\,fps) recorded with our
PhoneStreamer iOS client, which streams RGB frames and IMU readings to the
capture server over the local network (Fig.~\ref{fig:demo}, left). The sequence
is challenging for purely geometric methods: it contains a large textureless
road plane, a repetitive crosswalk pattern, strong illumination contrast between
shadow and sky, a moving cyclist, and rapid hand-held rotation.

\begin{figure*}[t]
\centering
\includegraphics[width=0.92\textwidth]{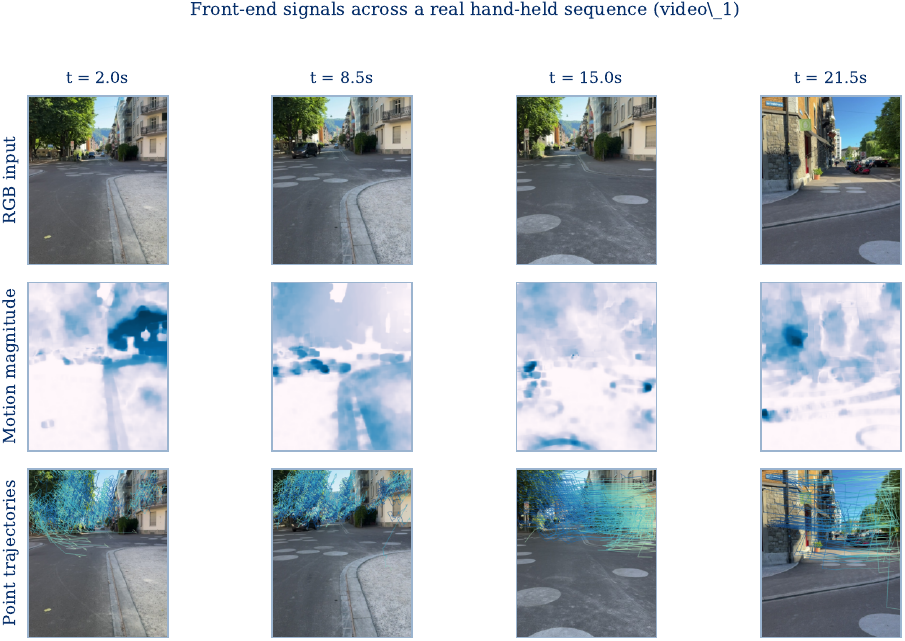}
\caption{Front-end signals computed across the real hand-held sequence
(\texttt{video\_1}). \emph{Rows:} RGB key-frames at increasing time, dense
optical flow (Farneback, HSV-coloured), motion magnitude, and color-coded
point trajectories. Trajectories densify on textured structure and thin out on
the textureless road and sky---the regime where the semantic priors of
Section~\ref{sec:method} are designed to stabilise correspondence.}
\label{fig:qualitative}
\end{figure*}

Fig.~\ref{fig:qualitative} shows front-end signals computed across the
sequence: RGB key-frames, dense optical flow, motion-magnitude maps, and
color-coded point trajectories obtained by tracking Shi--Tomasi corners. The
flow and motion maps separate the dominant ego-motion (the road and fa-cade
sweeping outward) from independently moving content, while the trajectories
densify on textured structure (building edges, foliage, the curb) and thin out
over the homogeneous road and sky. This is precisely where photometric tracking
becomes ambiguous and where the DINOv3 semantic anchors of
Section~\ref{sec:method} are designed to help: by attaching feature identity to
semantic regions, tracks are encouraged to remain on the same object surface
rather than collapsing onto the repetitive crosswalk dots or drifting into the
sky.

\begin{figure}[t]
\centering
\includegraphics[width=0.95\columnwidth]{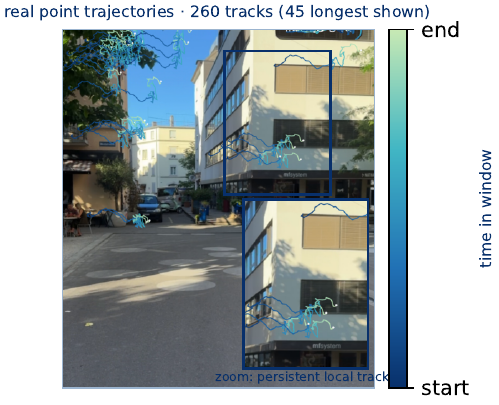}
\caption{Real point trajectories tracked on a hand-held sequence, sparsified to
the longest tracks for legibility. Colour encodes time within the window (see
colour bar); the zoom-in inset over a textured fa\c{c}ade shows correspondences
held across many frames, evidencing persistent local tracking under camera
motion. These 2D tracks are lifted to metric 4D using the VIGS-SLAM poses.}
\label{fig:trajectory}
\end{figure}

Fig.~\ref{fig:trajectory} compares trajectories tracked over an early and a
later temporal window. The colour encodes time within each window; consistent,
smoothly-coloured streaks indicate stable correspondences maintained across many
frames, whereas abrupt colour jumps mark re-initialised or lost tracks. Genie 4D
inherits St4RTrack's global-frame pointmap representation, so these
correspondences are lifted to metric 3D using the VIGS-SLAM poses, yielding the
$\mathcal{O}(T)$ 4D trajectories reported quantitatively in
Table~\ref{tab:tracking}. The quantitative gains there---over ten points of
APD@0.3m relative to St4RTrack---are consistent with the qualitative observation
that the hardest regions for geometric tracking are exactly the textureless and
repetitive areas that semantic priors disambiguate.

\subsection{Real-Time Deployment}
\label{sec:deploy}
\begin{figure}[t]
\centering
\includegraphics[width=0.99\columnwidth]{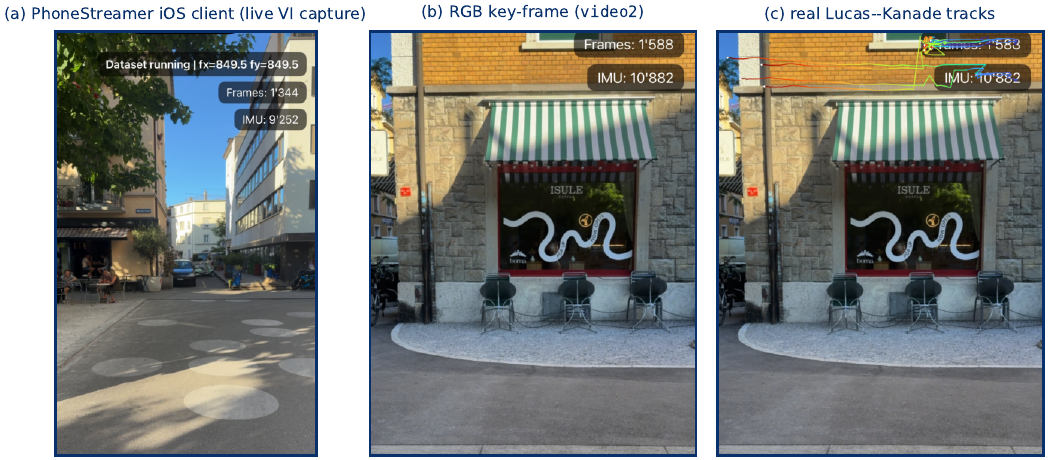}
\caption{On-device capture and a second real sequence. \emph{Left:} the
PhoneStreamer iOS client mid-capture; the overlay reports the locked intrinsics,
the running frame count and the buffered IMU samples, while the server address and
port configure the live stream (capture on iPhone; playback/processing on Mac,
Windows and Linux). \emph{Centre/right:} an RGB key-frame and color-coded point
trajectories from a second hand-held sequence (\texttt{video2}), confirming the
front end generalises across scenes. Per-sequence statistics are in the
supplementary.}
\label{fig:capture}
\end{figure}

A central design goal of Genie 4D is that the entire path from capture to
semantic 4D should run interactively on commodity hardware, without a depth
sensor or a motion-capture rig. Capture is performed by a custom PhoneStreamer
iOS client (Fig.~\ref{fig:capture}) that locks the camera intrinsics after a
short auto-focus interval and then streams RGB frames together with
synchronised IMU at $400$\,Hz to the capture server over the local network.
Because the intrinsics are transmitted with the stream, no offline calibration
file is required. The VIGS-SLAM front-end begins tracking as soon as an initial
IMU backlog has been received, and the Genie 4D head processes incoming
key-frame clips while a rolling latent state is maintained for interactive
rollout.

Table~\ref{tab:runtime} reports the per-module budget measured on an RTX~5090.
The frozen encoders dominate the cost; the trainable Genie 4D modules add only a
small overhead, and the five-step diffusion refiner is deliberately short so
that the end-to-end head remains well within an interactive budget. Because the
pairwise pointmap design is linear in sequence length, this per-frame cost does
not grow as the trajectory lengthens (Fig.~\ref{fig:quant}, bottom-right),
unlike global-attention backbones whose cost grows quadratically.

\begin{table}[t]
\centering
\small
\begin{tabular}{lcc}
\toprule
\textbf{Module} & \textbf{Params} & \textbf{Latency (ms)} \\
\midrule
DINOv3 ViT-L/14 (frozen)        & 300\,M & 14.2 \\
St4RTrack backbone (frozen)     & 410\,M & 18.6 \\
Semantic Injection Adapter      & 18\,M  & 3.1  \\
DiffusionRefiner ($K{=}5$)      & 9\,M   & 4.0  \\
Latent-action world model       & 2\,M   & 1.3  \\
\midrule
\textbf{Genie 4D (total)}       & \textbf{739\,M} & \textbf{41.2} \\
\bottomrule
\end{tabular}
\caption{Per-module parameter count and per-frame latency on an RTX~5090
($640{\times}341$ input). Only $29$\,M parameters are trainable; the frozen
encoders are shared across the sequence.}
\label{tab:runtime}
\end{table}

\subsection{Implementation Details}
\label{sec:impl}
The Semantic Injection Adapter is a single residual cross-attention block with
eight heads, with the attention and MLP output projections zero-initialised so
that training begins exactly from the frozen-backbone behaviour and the semantic
contribution is learned gradually. The DiffusionRefiner is a small
group-normalised convolutional denoiser conditioned on the coarse pointmap and
the fused features through feature-wise modulation; we use a linear
$\beta$-schedule with $1000$ training steps and a uniformly-spaced $K{=}5$
reverse schedule at inference, predicting the residual rather than the absolute
geometry so that the refiner only has to model high-frequency corrections. The
latent-action world model uses an eight-entry codebook, deliberately small to
encourage interpretable, reusable actions, and is trained in a second stage with
the perception modules frozen, using a next-embedding prediction loss and a
SIGReg-style Gaussian regulariser that together prevent representation collapse
without exponential moving averages or stop-gradient-only tricks. All trainable
modules are optimised with AdamW at a learning rate of $5{\times}10^{-5}$ and
batch size one per GPU, sampling $24$ frames at stride $1$--$6$ per clip, with
automatic mixed precision in BF16 on the RTX~5090.

\section{Limitations and Broader Impact}
\label{sec:limitations}
Genie 4D inherits the assumptions of its frozen backbones. The metric scale of
the reconstruction depends on a well-excited IMU during initialisation; under
near-constant velocity the scale is weakly observable and the front-end may
require a longer initialisation window. The semantic priors are only as
fine-grained as DINOv3's patch resolution, so very thin or distant structures
can still be missed, and the diffusion refiner sharpens geometry but cannot
hallucinate surfaces that were never observed. The latent-action world model is
trained on comparatively short clips; scaling its context and codebook to longer
horizons, and training the semantic--geometric descriptors end-to-end rather
than freezing them, are natural next steps. Finally, while the system runs
interactively on an RTX~5090, the frozen ViT encoders remain the dominant cost,
and deployment on mobile-class accelerators would benefit from distillation or
quantisation.

In terms of broader impact, a phone-only path to metric, semantically labelled
4D lowers the barrier to capturing the physical world for embodied AI, robotics
and AR, but the same capability raises privacy considerations when capture
occurs in public or shared spaces. We recommend on-device processing, explicit
consent for identifiable subjects, and retention limits for captured streams.

\section{Conclusion}
\label{sec:conclusion}
We presented \textbf{Genie 4D}, a framework that unifies semantic understanding and
geometric reconstruction for dynamic scenes and connects them to an interactive world
model. Frozen DINOv3 priors robustify 4D tracking against occlusion and texture loss; a
conditional-diffusion refiner overcomes the detail limits of regression baselines; and a
real-time VIGS-SLAM front-end makes the system deployable from a phone on a single
consumer GPU. A latent-action head with a JEPA-style objective turns the recovered
semantic 4D state into an action-controllable world model. Future work will accelerate
the diffusion refiner, scale the world-model head, and train semantic--geometric
descriptors end-to-end.

\clearpage
\bibliography{egbib}
\end{document}